\def\paperID{5323}
\def\confName{ICCV}
\def\confYear{2025}

\def\shortName{visual action prompts}  %
\def\paperTitle{Precise Action-to-Video Generation Through Visual Action Prompts}

\def\authorBlock{
    Yuang Wang$^{2^{*}}$ \quad
    Chao Wen$^{3^{*}}$ \quad
    Haoyu Guo$^{2^{*}}$ \quad
    Sida Peng$^{2}$ \quad
    Minghan Qin$^{4}$ \\
    Hujun Bao$^{2}$ \quad
    Xiaowei Zhou$^{2}$ \quad
    Ruizhen Hu$^{1,5^{\dagger}}$ \\[2.0mm]
    $^{1}$Xiangjiang Lab \quad 
    $^{2}$Zhejiang University \quad
    $^{3}$Fudan University \\
    $^{4}$Tsinghua University \quad
    $^{5}$Shenzhen University
}

\newif\ifreview 
\newif\ifarxiv \newcommand{\arxiv}{\arxivtrue}
\newif\ifcamera 
\newif\ifrebuttal 

\arxiv %

\pdfoutput=1
\documentclass[10pt,twocolumn,letterpaper,table,xcdraw]{article}
\ifreview \usepackage[review]{cvpr} \fi
\ifarxiv \usepackage[pagenumbers]{cvpr} \fi
\ifrebuttal \usepackage[rebuttal]{cvpr} \fi
\ifcamera \usepackage{cvpr} \fi

\usepackage{graphicx}	
\usepackage{amsmath}	
\usepackage{amssymb}	
\usepackage{booktabs}
\usepackage{makecell}
\usepackage{times}
\usepackage{microtype}
\usepackage{epsfig}
\usepackage{caption}
\usepackage{float}
\usepackage{placeins}
\usepackage{color, colortbl}
\usepackage{stfloats}
\usepackage{enumitem}
\usepackage{tabularx}
\usepackage{xstring}
\usepackage{multirow}
\usepackage{xspace}
\usepackage{url}
\usepackage{subcaption}
\usepackage{lipsum}
\usepackage{blindtext}
\usepackage[hang,flushmargin]{footmisc}
\usepackage{mwe}
\usepackage{algorithm}
\usepackage{algpseudocode}
\usepackage{nicefrac}
\usepackage{suffix}

\ifcamera \usepackage[accsupp]{axessibility} \fi

\newcommand{\nbf}[1]{{\noindent \textbf{#1.}}}
\WithSuffix\newcommand\nbf*[1]{{\noindent \textbf{#1}}}

\newcommand{\supp}{supplemental material\xspace}
\ifarxiv \renewcommand{\supp}{appendix\xspace} \fi

\definecolor{purple}{rgb}{0.65,0,0.65}

\definecolor{color1}{RGB}{249, 102, 102}
\definecolor{color2}{HTML}{579BB1}
\definecolor{color3}{RGB}{104, 185, 132}

\newcommand{\R}[1]{{%
      \textbf{%
        \ifstrequal{#1}{1}{\textcolor{color1}{R#1}}{%
          \ifstrequal{#1}{2}{\textcolor{color2}{R#1}}{%
            \ifstrequal{#1}{3}{\textcolor{color3}{R#1}}{%
              \ifstrequal{#1}{4}{\textcolor{teal}{R#1}}{%
                \textcolor{cyan}{R#1}%
              }}}}%
      }%
    }}
\WithSuffix\newcommand\R*[1]{{%
      \textbf{%
        \ifstrequal{#1}{1}{\textcolor{color1}{aG17}}{%
          \ifstrequal{#1}{2}{\textcolor{color2}{c51B}}{%
            \ifstrequal{#1}{3}{\textcolor{color3}{R5UR}}{%
              \ifstrequal{#1}{4}{\textcolor{teal}{R#1}}{%
                \textcolor{cyan}{R#1}%
              }}}}%
      }%
    }}
\newcounter{CQ} 
\newcounter{QR1} 
\newcounter{QR2} 
\newcounter{QR3} 
\newcommand{\printcounter}[1]{\arabic{#1}\stepcounter{#1}}
\newcommand{\Q}[2]{{%
      \noindent
      \textbf{%
        \ifstrequal{#1}{0}{{CQ.\printcounter{CQ} - #2.}}{%
          \ifstrequal{#1}{1}{{\textcolor{color1}{Q.\printcounter{QR1}} - #2.}}{%
            \ifstrequal{#1}{2}{{\textcolor{color2}{Q.\printcounter{QR2}} - #2.}}{%
              \ifstrequal{#1}{3}{{\textcolor{color3}{Q.\printcounter{QR3}} - #2.}}{%
              }}}}%
      }%
    }}

\newcommand\blfootnote[1]{%
  \begingroup
  \renewcommand\thefootnote{}\footnote{#1}%
  \addtocounter{footnote}{-1}%
  \endgroup
}

\usepackage{xr-hyper}

\makeatletter
\newcommand*{\addFileDependency}[1]{
  \typeout{(#1)}
  \@addtofilelist{#1}
  \IfFileExists{#1}{}{\typeout{No file #1.}}
}

\makeatother

\definecolor{cvprblue}{rgb}{0.21,0.49,0.74}
\usepackage[pagebackref,breaklinks,colorlinks,allcolors=cvprblue]{hyperref}
\usepackage[capitalize]{cleveref}
\crefname{section}{Sec.}{Secs.}
\crefname{table}{Table}{Tables}
\crefname{figure}{Fig.}{Figs.}

\ifarxiv \crefname{appendix}{App.}{Apps.}
\else \crefname{appendix}{Suppl.}{Suppls.} \fi

\begin{document}
\title{\paperTitle}
\author{\authorBlock}
\twocolumn[{%
    \centering
    \vspace{-19pt}
    \renewcommand\twocolumn[1][]{#1}%
    \maketitle
    \includegraphics[width=0.9\linewidth]{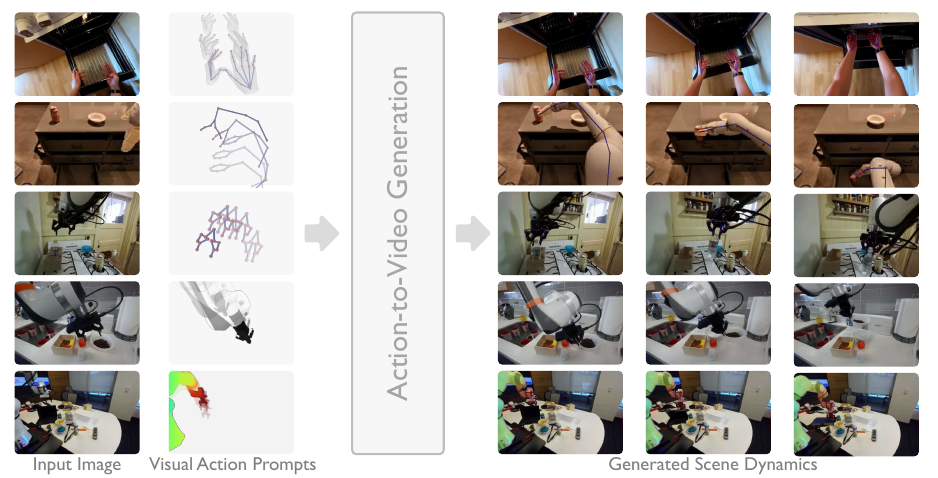}
    \captionof{figure}{We propose visual action prompts as unified representations for complex, high-degree-of-freedom actions (\eg, simulating scene dynamics driven by human hands or robotic grippers). Visual action prompts are ``renderings'' of subjects' action-induced 3D strucutres, among which we use skeleton as the primary representaion for its acquisition efficiency. This paradigm enables training action-driven video generation models across heterogeneous datasets while facilitating cross-domain knowledge transfer. \vspace{1em}}
    \label{fig:teaser}
}]

\blfootnote{$^*$Equal contribution. $^\dagger$Corresponding author: Ruizhen Hu.}

\begin{abstract}
We present \shortName, a unified action representation for action-to-video generation of complex high-DoF interactions while maintaining transferable visual dynamics across domains.
Action-driven video generation faces a precision-generality tradeoff: existing methods using text, primitive actions, or coarse masks offer generality but lack precision, while agent-centric action signals provide precision at the cost of cross-domain transferability.
To balance action precision and dynamic transferability, we propose to ``render'' actions into precise visual prompts as domain-agnostic representations that preserve both geometric precision and cross-domain adaptability for complex actions; specifically, we choose visual skeletons for their generality and accessibility.  
We propose robust pipelines to construct skeletons from two interaction-rich data sources --- human-object interactions (HOI) and dexterous robotic manipulation --- enabling cross-domain training of action-driven generative models.
By integrating visual skeletons into pretrained video generation models via lightweight fine-tuning, we enable precise action control of complex interaction while preserving the learning of cross-domain dynamics.
Experiments on EgoVid~\cite{wang2024egovid}, RT-1~\cite{rt1} and DROID~\cite{droid} demonstrate the effectiveness of our proposed approach.
\end{abstract}

\vspace{-1em}
\section{Introduction}
\label{sec:intro}

With improvements in quality and controllability of visual generative models~\cite{cogvideox,openai_sora,bar2024lumiere,polyak2025moviegen}, action-driven generative models are now widely applied in gaming~\cite{genie,gamengen,gamegenx,matrix,gamefactory}, decision-making~\cite{unipi,unisim}, robot learning and simulation~\citep{ivideogpt,diamond,irasim,cosmos}. These frameworks utilize action sequences as conditional inputs to video generation models, producing video frames that depict the outcomes of those actions.
This paper focuses on action-driven generative models under complex, high-DoF action control, such as simulation of scene dynamics governed by human hands and robotic grippers operations.

The primary challenge lies in the absence of a unified yet precise action representation to effectively model high-DoF and heterogeneous actions across diverse fields and applications, impeding the training of a unified model facilitating knowledge transfer across domains.
Different action representations have been proposed in diverse domains, such as text~\cite{unisim}, high-level primitive skills~\cite{genie,gao2025vista,gamengen,gamegenx}, and low-level states of specific agent configurations~\cite{ivideogpt,irasim,cosmos}. 
However, action representations face a precision-generality trade-off. 
Text, though universally applicable, can only present the high-level intention of an action.
Pre-defined skills like character movement and gaming actions (\eg, shooting~\cite{gamegenx,diamond}) are similar to text, both of which present high-level action primitives, while general, fail to represent complex, low-level character motions and intricate interactions with the environment that are critical in applications like motion sensing gaming.
The low-level state of specific agent configuration (\eg, a robot arm's end-effector 6D pose and gripper openness) is an almost lossless action representation.  %
It is adopted in domains requiring the utmost precision like robotic simulation and planning~\cite{ivideogpt,irasim,cosmos}. However, it tightly couples the action signal to specific embodiments, lacking generality.

To achieve a balanced precision-generality tradeoff, we propose using \emph{precise visual action prompts} as unified control signals for interactive generative models driven by complex actions, while retaining generality across domains.
Visual action prompts are produced by ``rendering'' the 3D structure of actions-induced agent states into image space, which can be in different forms, such as coarse masks, colored renderings or depth maps, and 2D skeletons.
They can effectively represent actions of high-DoF subjects --- such as human hands, robot grippers, and dexterous hands --- with high precision.

Some forms of visual action prompts come with notable drawbacks. Mesh-rendering-based approaches require reconstruction of complete meshes, which are challenging to scale up with in-the-wild data. Coarse subject masks~\cite{coshand,akkerman2024interdyn} are easier to recover but suffer from occlusions; their limited precision is also problematic in fine-grained tasks like robot simulation. To balance the ease of recovery and action precision, we adopt \emph{skeletons} as the unified control signal, which have long been a universal tool in animation~\cite{parent2012computer,lewis2023pose,baran2007automatic,mourot2022survey,hu2024animate}.
They can be robustly recovered from in-the-wild data~\cite{easymocap,shen2024world,dong2020motion,horaud1995hand,strobl2006optimal,tsai1989new}, facilitating large-scale training across domains.

To demonstrate the effectiveness of visual action prompts, we propose scalable strategies to recover complete skeletons of human hands and robot grippers on datasets including EgoVid~\cite{wang2024egovid}, RT-1~\cite{rt1} and DROID~\cite{droid}. Then, we finetune a base video generation model~\cite{cogvideox} to adapt it to an action-controllable model supporting intricate interaction.
We further show that visual action prompts serve as a more precise and easier to learn control signal of action for video models, compared to text and agent-centric states. Moreover, we show that visual action prompts enable training a unified model on multi-domain data including HOI and robot manipulation, which facilitates cross-domain knowledge transfer of interaction-driven dynamics. 

In summary, we make the following contributions:
\begin{itemize}
\item We propose using precise visual action prompts, specifically skeletons, as the unified action representation for action-driven generative models in scenarios involving complex, high-DoF actions.
\item We introduce scalable strategies to recover skeleton-based visual action prompts on interaction-rich datasets including HOI and robot manipulation.
\item We demonstrate visual action prompts' advantages including ease of learning, precision and generality, which enable joint training on heterogeneous data and facilitate knowledge transfer.
\end{itemize}

\section{Related Work}
\label{sec:related}

\begin{figure*}[!t]
    \centering
    \includegraphics[width=\linewidth]{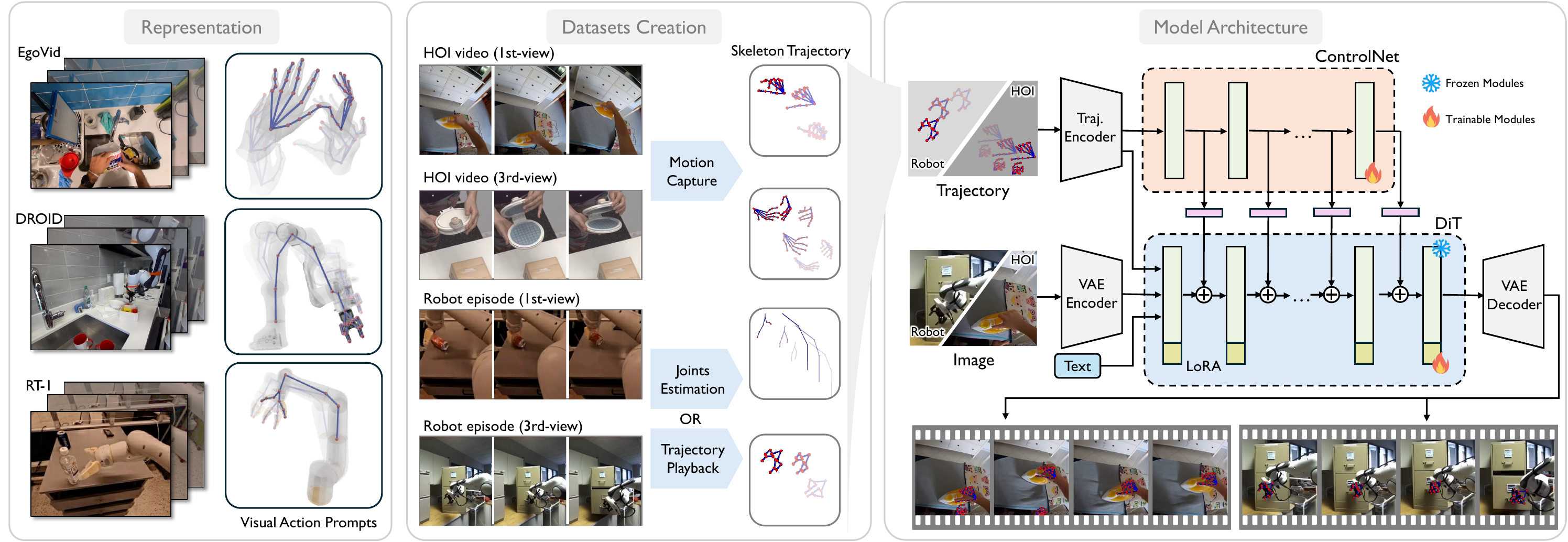}
    \vspace{-15pt}
    \caption{\textbf{Action-to-video generation with visual action prompts.} 
    We project action-induced 3D structural dynamics of diverse agents into 2D visual action prompts, primarily 2D skeletons, establishing a unified control signal for action-conditioned video generation. 
    We design data creation pipelines for HOI and robot videos to robustly recover their 2D skeletons. The constructed visual action prompts are injected into a pretrained video generation model for fine-tuning and generating plausible interaction-driven visual dynamics.
    }
    \vspace{-1em}
    \label{fig:pipeline}
\end{figure*}

\nbf{Action-to-video generation}
Recent works have been pursuing action control of video models to enable interaction-rich applications like gaming and agent training~\cite{genie,diamond,deepmind2024genie2,gamengen,gamegenx,gamefactory}, robotic simulation/learning~\cite{irasim,ivideogpt,cosmos}, and general decision making~\cite{yang2024VideoAsNewLang}.
Typical game actions involve predefined primitives where state transitions follow explicit rules or rules combined with physics simulation. Video-driven game research explores various action representations: Genie~\cite{genie} learns a discrete set of latent actions from unlabeled videos to drive generation; GameNGen~\cite{gamengen} and DIAMOND~\cite{diamond} directly map primitive actions (e.g., directional moves, shooting) to video frames. Recent efforts further scale these approaches~\cite{oasis,gamefactory,gamegenx,matrix},
yet remain constrained by primitive action representations that limit complex environmental interactions.
Video generation for world model and RL agents training~\cite{diamond,ivideogpt,genie,deepmind2024genie2} are also still limited to learning predefined primitive actions, incapable of simulating high-DoF embodiments
interacting with complex environments.
In simulation, UniSim~\cite{unisim} represents high-level action intentions with texts, which is general but lacks precision. Text as action is more plausible for general decision making like generating how-to guides~\cite{GenHowTo} and high-level policies for robotics~\cite{unipi,ko2023learning,susie,bharadhwaj2024gen2act}. In the other end, IRASim~\cite{irasim} and Cosmos~\cite{cosmos} employ agent-centric actions like 7-DoF end-effector states, which is almost lossless for positional control but not general. 
Recently, CosHand~\cite{coshand} and InterDyn~\cite{akkerman2024interdyn} explore generating future state images or videos using coarse hand masks as indications of actions.
However, mask-based representations are fragile to occlusions, and segmented masks are often imprecise, which hinders applications requiring highly accurate actions.

\nbf{Motion control and character animation}
Controllability of video generation models is critical for downstream applications. Extensive research focuses on achieving control over the synthesized content in different levels, especially for the target agent.
Region-based methods~\cite{wang2024boximator,shi2024motioni2v} provide high-level guidance for local motion. Sparse/dense trajectory-based methods~\cite{wang2024motionctrl,yin2023dragnuwa,zhang2024tora,geng2024motion} govern local object movements or camera motion. Character animation techniques further enable precise control through skeletons~\cite{hu2024animate,zhang2024mimicmotion,men2024mimo}, mesh renderings~\cite{zhu2024champ,zhou2024realisdance}, or reference videos~\cite{wang2024motioninversion,hu2025animate2}, all requiring explicit dynamic specifications for the control targets.
In contrast, we aim to generate interaction-induced dynamics of the whole environment by providing the model with action signals only, without relying on its inherent dynamics. 

\nbf{Dynamic-rich datasets}
Effectively orchestrating diverse datasets with rich interaction and interaction-driven dynamics lays the groundwork for action-driven video generation.
Video datasets about human-object-interaction (HOI) are natural sources for learning interaction-driven dynamics.
Early works like SSV2 and Kinetics~\cite{goyal2017something,kay2017kinetics} collects datasets of humans performing basic actions with everyday objects. Despite their substantial scale, their relatively low video quality falls short of the standards required by modern video generation models.
Recent advancements have introduced large-scale egocentric human activity datasets~\cite{damen2018scaling,grauman2022ego4d,grauman2024ego,long2024babyview} to advance behavioral understanding and learning. A notable example, Ego4D, offers 3,670 hours of daily activity videos across diverse scenarios. However, its lengthy sequences are suboptimal for generation tasks. EgoVid-5M~\cite{wang2024egovid} addresses this by curating trimmed, filtered, and captioned clips from Ego4D, making them more plausible for generative models.
Beyond video data, specialized datasets for HOI motion research provide 3D hand-object annotations~\cite{liu2024taco,fan2023arctic,oakink2,banerjee2024hot3d,fu2024gigahands}. While valuable for high-precision 3D-controlled fine-tuning, their limited diversity restricts utility in foundational interaction-driven dynamics learning.
Complementing HOI resources, embodied AI and robotics research has yielded high-quality interaction data as well. Open X-Embodiment~\cite{oxe} aggregates multi-task datasets of complex robotic interactions across embodiments. We select RT-1~\cite{rt1} and DROID~\cite{droid} from this collection for their scale and relatively precise camera calibration.

\setlength{\abovedisplayskip}{4pt} 
\setlength{\belowdisplayskip}{4pt}
\setlength{\abovedisplayshortskip}{4pt}
\setlength{\belowdisplayshortskip}{4pt}

\section{Method}
\label{sec:method}

Given an image observation as the initial frame and a sequence of actions from human hands or robot grippers, our goal is to generate videos that accurately depict the interaction outcomes under precise action control. To achieve this, we introduce a general and precise visual action prompt for video generation models.
\cref{fig:pipeline} illustrates our framework, which includes visual action representation, dataset construction, and visual dynamic modeling. 

To maximize data scale and interaction relevance, we focus on two primary agents: human hands and robotic grippers. 
Their actions --- despite kinematic differences --- are uniformly encoded as skeletons, as our visual prompts (\cref{subsec:problem_formulation}). 
To train our model, we process and annotate two types of datasets (\cref{subsec:dset_creation}): (1) human hand skeletons extracted via motion capture from HOI videos; and (2) robotic gripper skeletons are synthesized through joint-state rendering from robotic episodes.
Leveraging these large-scale (skeleton, video) pairs, we fine-tune video generation models to enable visual action prompt control (\cref{subsec:architecture}).

\subsection{Visual action prompts}
\label{subsec:problem_formulation}

Our goal is to develop a generalizable video generation model capable of synthesizing plausible scene dynamics and interaction outcomes $\mathbf{s}_{1:t} \in \mathcal{S}$, given the initial observation $\mathbf{s}_0$ and driven by user-specified complex action sequences $\mathbf{a}_{0:t-1} \in \mathcal{A}$. 
The problem can be formulated as learning the conditional probability distribution $\mathcal{P}$ governing the state representation:  
\begin{equation}
    \mathbf{s}_{1:t} \sim P(\mathbf{s}_{1:t} | \mathbf{s}_0, \mathbf{a}_{0:t-1}).
\end{equation}

While the formulation is brief, there are certain challenges in practice: 
(1) accommodating diverse configurations of intricate action spaces while ensuring compatibility across tasks, and (2) retaining the transferability of visual dynamics under precise action control, thus enabling the model to learn from scalable datasets composed of diverse domains.

To resolve those challenges, our core insight is to map the action sequence $\mathbf{a}_{0:t-1}$ to \emph{visual action prompts} $\mathbf{v}_{1:t}$ as follows:
\begin{equation}
    \mathbf{v}_{1:t} = \mathcal{R}(\mathbf{a}_{0:t-1}) \in \mathbb{R}^{T \times H \times W \times C},
\end{equation}
where $T$ represents the action trajectory length, $H$ and $W$ represent the image height and width, $C$ is the number of channels, determined by the specific visual representation, and $\mathcal{R}$ indicates rendering operation according to known camera parameters.
We consider mesh-based renderings (e.g., colored images, depth maps) and 2D skeletons as \emph{precise} visual action prompts.
Given the challenges associated with recovering fine-grained meshes from in-the-wild data, we opt to employ 2D skeletons as our primary representation.

\subsection{Scalable dataset creation}
\label{subsec:dset_creation}

\begin{figure}[t]
    \centering
    {
    \captionsetup[subfigure]{skip=-2pt}
    \begin{subfigure}{\linewidth}
        \centering
        \includegraphics[width=\linewidth]{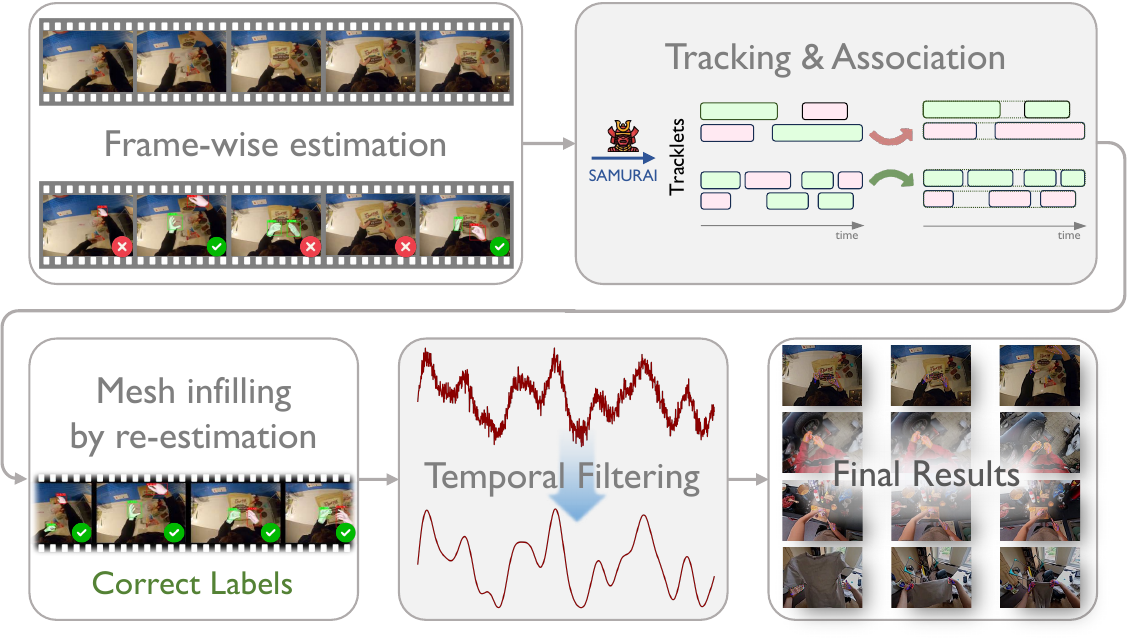}
        \caption{HOI dataset creation pipeline.}
        \label{fig:hoi_creation}
    \end{subfigure}
    \vskip 4mm
    \begin{subfigure}{\linewidth}
        \centering
        \includegraphics[width=\linewidth]{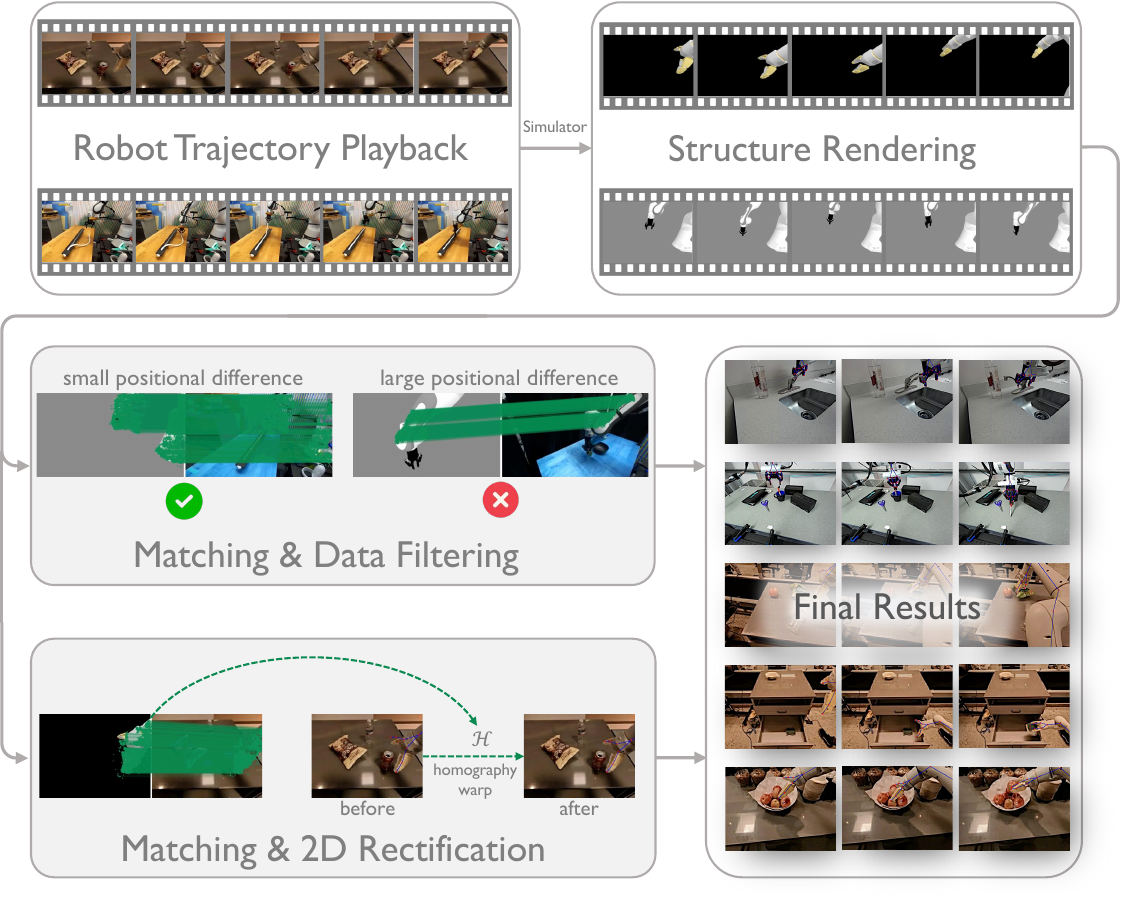}
        \caption{Robot dataset creation pipeline.}
        \label{fig:robot_creation}
    \end{subfigure}
    }
    \vspace{-10pt}
    \setlength{\belowcaptionskip}{-2pt}
    \caption{Pipelines of recovering 2D skeletons for visual action prompts from human-object interaction (HOI) and robotic datasets.}
    \label{fig:hoi_dset}
\end{figure}

To enable the learning of transferable visual dynamics under precise action conditioning, we construct large-scale (skeleton, video) pairs from two distinct sources: skeletons are estimated from in-the-wild HOI videos via our proposed pipeline (\cref{fig:hoi_creation}), while for robot manipulation episodes, they are rendered directly from state logs, followed by an optional correction step to ensure tight alignment with the visual observations (\cref{fig:robot_creation}).

\nbf{In-the-wild HOI videos}
We leverage in-the-wild HOI-centric videos for their rich hand-object interactions and scene dynamics, which is ideal for learning visual dynamic models. However, severe occlusions in these videos make direct 2D pose estimation unreliable~\cite{lugaresi2019mediapipe,yang2023dwpose,khirodkar2024sapiens}. To robustly extract 3D hand mesh trajectories, we introduce a four-stage pipeline that addresses common failures like missed detections and temporal jitter (see~\cref{alg:tracking_association} and~\cref{fig:hoi_creation}): (1) Initialization: We first detect all potential hands per-frame using 3d hand mesh recovery method Wilor~\cite{wilor}. (2) Temporal Stabilization: We then form consistent tracklets and correct handedness errors using SAMURAI~\cite{samurai}. (3) Refinement: Missing meshes within these tracklets are re-estimated. (4) Smoothing: Finally, we apply a OneEuro filter~\cite{oneeuro} to the MANO trajectories to eliminate jitter.

\begin{algorithm}[!t]
	\caption{Tracking and Association}
	\label{alg:tracking_association}
	\begin{algorithmic}[1]
		\State $\mathcal{B} \gets \emptyset$                                \Comment{Initialize the bounding box set}
		\State $\mathcal{T} \gets \emptyset$                                \Comment{Initialize the tracklet set}
        \State $\mathcal{A} \gets \{\}$                                               \Comment{Initialize the association dictionary}
        \State 

		\ForAll{frame $F$ in video}
			\State $\mathcal{B} \gets \mathcal{B} \cup \text{Detect}(F)$          \Comment{Based on Wilor}
		\EndFor

        \State $\mathcal{B}_{untracked} \gets \text{Sort}(\mathcal{B})$               \Comment{By descending confidence}

		\While{$\mathcal{B}_{untracked} \neq \emptyset$}
            \State $B \gets \mathcal{B}_{untracked}.\text{pop}(0)$
			\If{$\max_{T \in \mathcal{T}} \text{IoU}(B,T) < \theta_{IoU}$}
                \State $T_{\text{new}} \gets \text{Track}(B)$    \Comment{Based on SAMURAI}
                \State $\mathcal{T}.\text{append}(T_{\text{new}})$
			\EndIf
		\EndWhile

        \ForAll{$(B,T) \in \mathcal{B}\times\mathcal{T}$}
            \If{$\text{IoU}(B,T) \geq \theta_{IoU}$}
                \State $\mathcal{A}[T].\text{append}(B)$
            \EndIf
        \EndFor
        
        \State $\mathcal{T} \gets \text{HandnessFilter}(\mathcal{T}, \mathcal{A})$
        \State $\mathcal{T} \gets \text{Merge}(\mathcal{T}, \mathcal{A})$
        \State $\mathcal{T} \gets \text{NumberOfHandsFilter}(\mathcal{T}, \mathcal{A})$
	\end{algorithmic}
\end{algorithm}

\nbf{Robot manipulation episodes}
We also utilize robotic manipulation datasets (DROID~\cite{droid}, RT-1~\cite{rt1}), which offer focused interactions and scene dynamics while simplifying 3D skeleton extraction via robot state logs. However, camera calibration errors and temporal drift are common issues. To ensure precise 2D skeleton alignment with video observations, we implement a vision-based correction pipeline (\cref{fig:robot_creation}): (1) Episode Filtering: We use MatchAnything~\cite{matchanything} to match rendered robot meshes against real observations, discarding episodes with significant matching coordinates discrepancy. (2) Homography Rectification: For the remaining episodes with camera drfit, we apply per-frame homography warping to adjust the initial skeleton renderings in 2D, also guided by image matching, ensuring precise alignment with real-world observations.

\subsection{Visual dynamics model with precise control}
\label{subsec:architecture}
We build our model based on CogVideoX~\cite{cogvideox}, which is a text-to-video generation model pretrained on large-scale (text, video) pairs and further fine-tuned with (text, initial frame, video) triplets to a (text, image)-to-video model. Its architecture includes: a pretrained text encoder, a video VAE, and a DiT~\cite{dit} model with FullAttention for spatio-temporal video tokens and text token processing. Due to the high data demands of training visual dynamics models from scratch, we leverage CogVideoX as a pretrained base model.

To integrate visual action prompts, as shown in~\cref{fig:pipeline}, we first encode the control signals. Specifically, for controls in the form of skeleton or mesh, we render them as sequences of RGB images $\mathbf{v}_{1:t} \in \mathbb{R}^{T \times H \times W \times C}$, where $C=3$. These sequences are then fed into an 3D convolutional trajectory encoder to latent states $\mathbf{s}_{1:\nicefrac{t}{4}} \in \mathbb{R}^{\frac{T}{4} \times \frac{H}{8} \times \frac{W}{8} \times 16}$. For depth control, we directly feed $\mathbf{v}_{1:t}$ with $C=1$ into the encoder with the same architecture.

Direct supervised fine-tuning of the video generation model pretrained on massive datasets may lead to overfitting or the loss of generalized knowledge. Therefore we leverage ControlNet~\cite{controlnet} to inject the visual action prompts. Specifically, we create trainable copies of the first 14 blocks of the pretrained DiT with zero-initialized linear layers, and inject visual action prompts $\mathbf{s}_{1:\nicefrac{t}{4}}$ in these blocks.
Moreover, following Wonderland~\cite{wonderland}, we adopt a dual-branch conditioning mechanism by injecting $\mathbf{s}_{1:\nicefrac{t}{4}}$ also in the main DiT, through merging video and action tokens, and fine-tune the DiT backbone with LoRA~\cite{lora}.

During training, we amplify loss values around hand~/~gripper regions to prioritize learning the interaction and its induced dynamics.
To mitigate dominance of self-motion over interaction dynamics in robot videos with lengthy tasks, we sample more clips around timestamps where gripper state changes.

\section{Experiment}
\label{sec:exp}

Our experiments aim to validate two core claims: (1) visual action prompts outperform alternative control signals, \eg, text or agent-centric raw actions~/~states (\cref{subsec:agent_specific}) in driving interaction-aware scene dynamics; (2) the generality of visual action prompts across agent configurations and the effect of joint training on diverse datasets (\cref{subsec:agent_agnostic}). 
Moreover, ablation studies (\cref{subsec:ablation}) demonstrate the effectiveness of our model design and present results of different visual action prompts.

\nbf{Implementation details}
We curate three datasets with distinct characteristics.
\textbf{EgoVid}~\cite{wang2024egovid}: A subset of 200k training clips (from 1M clips) containing around 120 frames, 30 fps videos of diverse daily activities with hand skeletons. Clips with significant viewpoint changes are filtered via optical flow~\cite{teed2020raft} and point tracking~\cite{karaev2024cotracker3}. We manually select 32 clips including direct/indirect dynamics for evaluation.
\textbf{DROID}~\cite{droid}: A subset of 47k training clips of random third-person perspectives (from 76k episodes collected across 13 labs), with one lab's data reserved for evaluation on novel scenes. Tasks related to cleaning are retained for analyzing of novel skills. A total of 234 clips are used for quantitative assessment.
\textbf{RT-1}~\cite{rt1}: A subset of 57k training clips from 6 basic skills, two skills (``close drawer'' and ``place object upright'') are held out for evaluation. Notably, unlike previous works such as IRASim~\cite{irasim}, which focuses only on simulation of in-domain skills and scenes, we emphasize evaluating the interaction-driven dynamics of novel skills.
We caption all video clips with scene-centric captions via Qwen2.5-VL~\cite{bai2025qwen25vl}, only including scene arrangements and appearances while excluding action/dynamic descriptions. For training text-as-action models, we regenerate captions with explicit action annotations.
During training, we resize all video clips to the resolution of $720 \times 480$ and subsample video frames to 25 with variable fps in a plausible range.

\nbf{Metrics}
We utilize multiple metrics to evaluate the generated videos. To evaluate the \textit{visual similarity} between generated and ground truth videos, we report PSNR, SSIM~\cite{ssim}, and LPIPS~\cite{lpips}. To evaluate \textit{visual quality and temporal consistency}, we report the Fr\'{e}chet Video Distance (FVD)~\cite{fvd}. Finally, to explicitly evaluate \textit{dynamic correctness} of action and its impact on scene dynamics, we report the Spatio-temporal IoU proposed in Physics-IQ~\cite{physicsiq}.

\begin{figure}[tp]
    \centering
    \includegraphics[width=1.0\linewidth]{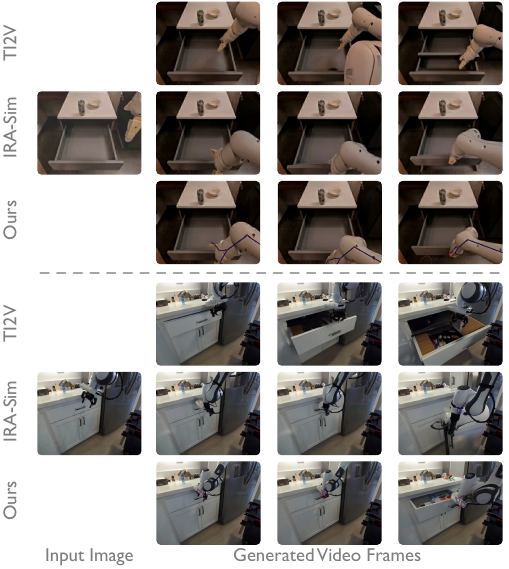}
    \caption{
        \textbf{Comparison of different action control signals.}
        Text as action (TI2V~\cite{cogvideox}) leads to ambiguity.
        Raw 7-DoF states/actions (IRASim~\cite{irasim}) leads to inaccurate control.
        Visual action prompts (Ours) facilitate dynamic learning under precise control.
    }
    \label{fig:control_signal}
\end{figure}

\subsection{Agent-specific control}
\label{subsec:agent_specific}

We demonstrate the superiority of visual action prompts over text-based and agent-centric action representations in control fidelity, scene dynamics plausibility, and learning efficiency. We conduct comparisons on two robot manipulation datasets RT-1~\cite{rt1} and DROID~\cite{droid}.
For text-based control, we implement high-level action guidance by fine-tuning the (text,image)-to-video model CogVideoX~\cite{cogvideox}. For agent-centric control, IRASim~\cite{irasim} directly employs 7-DoF robot-specific actions (end-effector poses and gripper states) as action trajectories to drive video generation. 
For a fair comparison, we reimplement IRASim's adaLN-based action injection upon CogVideoX and finetune the base model with 7-DoF state trajectories. 
We present quantitative comparisons between this reimplementation and the pretrained IRASim model in \supp to demonstrate its effectiveness.

\begin{table}[t]
    \centering
    \resizebox{0.96\columnwidth}{!}{%
    \begin{tabular}{llccccc}
        \toprule
        Dataset & Action Repr. & PSNR $\uparrow$ & SSIM $\uparrow$ & LPIPS $\downarrow$ & FVD $\downarrow$ & ST-IoU $\uparrow$ \\
        \midrule
        \multirow{2}{*}{RT-1~\cite{rt1}} 
        & Text (CogVideoX~\cite{cogvideox}) & 18.87 & 0.761 & 0.241 & 642.3 & 0.267 \\
        & Raw State (IRA-Sim~\cite{irasim}) & 23.96 & 0.854 & 0.127 & 302.2 & 0.507 \\
        & Skeleton (Ours) & \textbf{25.98} & 0.859 & \textbf{0.110} & 288.6 & \textbf{0.604} \\
        & Skeleton (Ours unified) & 24.90 & 0.847 & 0.121 & \textbf{258.1} & 0.576 \\
        \midrule
        \multirow{2}{*}{DROID~\cite{droid}}
        & Text (CogVideoX) & 18.10 & 0.790 & 0.200 & 248.3 & 0.239 \\
        & Raw State (IRA-Sim) & 20.13 & 0.825 & 0.146 & 151.2 & 0.365 \\
        & Skeleton (Ours) & 21.26 & 0.834 & 0.132 & 141.8 & 0.450 \\
        & Skeleton (Ours unified) & \textbf{21.58} & \textbf{0.836} & \textbf{0.126} & \textbf{124.4} & \textbf{0.478} \\
        \midrule
        \multirow{2}{*}{EgoVid~\cite{wang2024egovid}} 
        & Text (CogVideoX) & 13.44 & 0.440 & 0.503 & 1638.6 & -- \\
        & Skeleton (Ours) & 14.71 & 0.482 & 0.430 & 1243.6 & -- \\
        & Skeleton (Ours unified) & \textbf{14.93} & \textbf{0.486} & \textbf{0.421} & \textbf{1142.3} & -- \\
        \bottomrule
    \end{tabular}%
    }
    \caption{\textbf{Quantitative comparison on different datasets.} Visual action prompts (Ours) consistently outperform text-specified actions and raw agent-centric states. Joint training on all three datasets with a unified model leads to improved or comparable results across datasets.}
    \label{tab:joint_training}
\end{table}

\begin{table}[t]
    \centering
    \resizebox{0.96\columnwidth}{!}{%
    \begin{tabular}{@{\extracolsep{\fill}}l c c c@{}}
        \toprule
        Action Repr. & Known Lab \& Skill & Novel Lab & Novel Skill \\
        \midrule
        & \multicolumn{3}{c}{\small Mask IoU~$\uparrow$ / Boundary IoU~$\uparrow$ / $\mathcal{J} \& \mathcal{F}~\uparrow$} \\
        \midrule
        \multicolumn{4}{@{}l}{\textbf{Single dataset}} \\ %
        Text (CogVideoX)   & 34.1\,/\,31.9\,/\,33.0 & 22.8\,/\,30.9\,/\,26.9 & 20.5\,/\,25.2\,/\,22.9 \\
        Raw State (IRASim) & 49.1\,/\,48.9\,/\,49.0 & 25.8\,/\,40.9\,/\,33.4 & 34.9\,/\,40.6\,/\,37.8 \\
        Skeleton (Ours)    & \textbf{53.5\,/\,53.6\,/\,53.6} & \textbf{43.8\,/\,61.2\,/\,52.5} & \textbf{47.4\,/\,55.0\,/\,51.2} \\
        \midrule
        \multicolumn{4}{@{}l}{\textbf{Joint training}} \\
        Skeleton (Ours)    & \textbf{58.9\,/\,60.4\,/\,60.0} & \textbf{46.5\,/\,63.3\,/\,54.9} & \textbf{49.9\,/\,57.9\,/\,53.9} \\
        \bottomrule
    \end{tabular}%
    }
    \caption{\textbf{Quantitative comparison on different subset of the DROID~\cite{droid} dataset.} The manipulated object is annotated with point prompts in the first frame and tracked with SAM 2~\cite{ravi2024sam2}. We report metrics between the masks extracted from the generated and ground truth videos.
    \label{tab:droid_sam2}}
\end{table}

As shown in~\cref{fig:control_signal}, visual action prompts achieve better action fidelity through direct and precise action rendering, while text descriptions suffer from action ambiguity and less plausible generation. 
Although IRASim achieves effective control on RT-1 with fixed viewpoints, it fails to provide precise control on DROID with random third-person perspectives.
Quantitative comparisons in~\cref{tab:joint_training} further confirm the comprehensive superiority of visual action prompts across all metrics.

Due to the small proportion of interactive foreground in robot manipulation data, photometric metrics fail to effectively highlight differences between methods in generating interaction-driven dynamics, we conduct additional dynamic-centric evaluations on DROID. We manually annotate point prompts for interacted objects in the first frame, ensuring SAM 2~\cite{ravi2024sam2} successfully tracks them in the ground truth videos. Using the same prompts, we apply SAM-2 on generated videos and compute the $\mathcal{J} \& \mathcal{F}$ metric between generated tracking masks and ground truth masks to assess the model's quality in generating dynamics for interacted objects. As shown in~\cref{tab:droid_sam2}, visual action prompts achieve significant improvements across different DROID subsets.%

\begin{figure*}[tp]
    \centering
    \includegraphics[width=\linewidth]{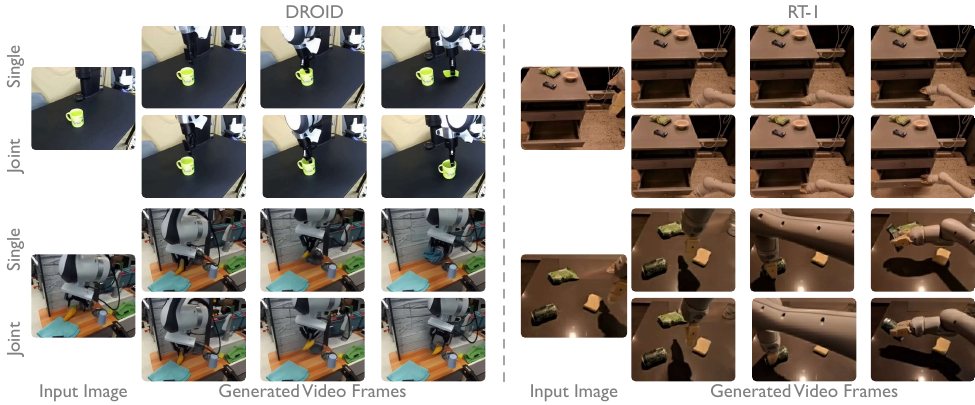}
    \caption{\textbf{Effect of joint training with visual action prompts.} Compared to single-dataset training, joint training leads to better object consistency on DROID~\cite{droid} and enables held-out skill execution (\eg, ``close the drawer'') on RT-1~\cite{rt1}.}
    \label{fig:joint_training}
\end{figure*}

\subsection{Agent-agnostic control}
\label{subsec:agent_agnostic}

Thanks to visual action prompts' balance over action precision and generality, we can integrate data from diverse domains of different agents to train a unified action-driven generative model. In this experiment, we demonstrate the effectiveness of visual action prompts for agent-agnostic control, where a single model drives the self motion and generates dynamics for agents with distinct configurations.

Specifically, we train a unified model on robotic datasets (RT-1~\cite{rt1}, DROID~\cite{droid}) and human egocentric video data (EgoVid~\cite{wang2024egovid}, based on Ego4D~\cite{grauman2022ego4d}). Conventional precise action representations struggle with multi-domain joint training: DROID and RT-1 feature incompatible robot configurations, and it is hard to map agent-centric signals like end-effector poses to DROID's random third-person camera viewpoints. EgoVid presents even greater challenges with dynamic egocentric views and complex human hand actions, which cannot be handled by previous action representations. %

\begin{figure}[tp]
    \centering
    \includegraphics[width=1.0\linewidth]{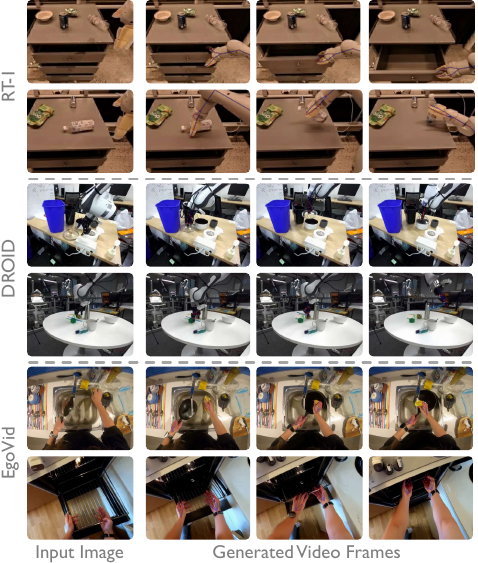}
    \caption{\textbf{Action-to-video generation of a unified model.}
    Visual action prompts enable joint training on diverse datasets and facilitate interaction-driven dynamic generation.
    The overlaid skeletons are only for visualization, demonstrating accurate action control.
    }
    \label{fig:more_qual_results}
\end{figure}

\begin{figure}[htbp]
    \centering
    \includegraphics[width=\linewidth]{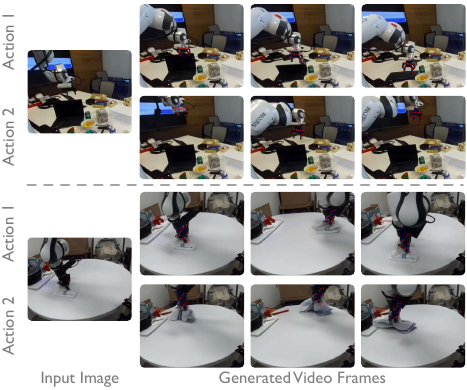}
    \caption{\textbf{Diverse generation results under different action trajectories.}
    Our model can simulate diverse actions and their visual outcomes from the same initial frame.
    }
    \label{fig:diverse_generation}
\end{figure}

As shown in~\cref{tab:joint_training}, joint training with visual action prompts achieves comparable or superior performance in generation quality and dynamic accuracy compared to single-domain training. Dynamic-centric evaluations on DROID in~\cref{tab:droid_sam2} further confirm its superiority.~\cref{fig:joint_training} highlights its specific advantages: joint training improves object consistency in DROID manipulations and enables novel skill generalization (\eg, closing drawers) on RT-1's unseen skill subset, which single-domain models fail to achieve. We attribute these benefits to visual action prompts simplifying the learning objective --- models bypass learning mapping from abstract action representations to agent motion and focus directly on learning interaction-driven dynamics induced by actions. We present more results of the unified model in~\cref{fig:more_qual_results}. We further illustrate in~\cref{fig:diverse_generation} that our model can generate diverse interaction outcomes aligned with different action trajectories under identical initial images, indicating its potential for downstream applications like simulation, planning, and robotic policy evaluation.

\begin{figure}[tp]
    \centering
    \includegraphics[width=\linewidth]{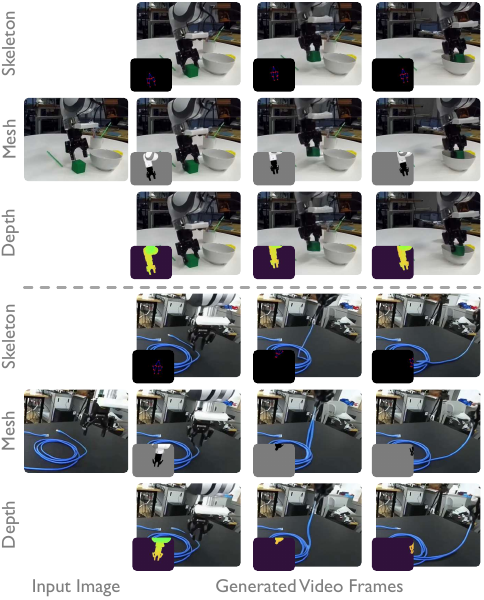}
    \caption{\textbf{Comparison of different forms of visual action prompts.}
    All three forms of visual action prompts can precisely represent delicate actions and drive plausible interactions.
    We prefer the skeleton-based prompt for its acquisition efficiency.
    }
    \label{fig:visual_prompts}
    \vspace{-5pt}
\end{figure}

\subsection{Ablation studies}
\label{subsec:ablation}

\nbf{Different forms of visual action prompts}
In~\cref{fig:visual_prompts} and~\cref{tab:ablation_visual_prompting}, we compare different visual action prompt forms: mesh rendering, depth maps, and our primary skeleton-based approach.
As shown in~\cref{tab:ablation_visual_prompting}, visual prompts with more details (mesh/depth) improve generation quality and dynamic accuracy compared to skeletons.
~\cref{fig:visual_prompts} demonstrates that all three forms effectively drive plausible interaction-aware dynamics. Given skeleton's lower acquisition cost for in-the-wild data and compatibility with additional sparse 3D information, we advocate its use as a unified action representation for large-scale training. For precision-critical applications, skeleton-driven models can be fine-tuned with higher-fidelity action prompts to fully leverage mixed-quality data.

\begin{table}[t]
    \centering
    \resizebox{0.96\columnwidth}{!}{%
    \begin{tabular}{lccccc}
        \toprule
        Control Method & PSNR $\uparrow$ & SSIM $\uparrow$ & LPIPS $\downarrow$ & FVD $\downarrow$ & ST-IoU $\uparrow$ \\
        \midrule
        Skeleton & 21.26 & 0.834 & 0.132 & 141.8 & 0.450 \\
        Mesh & \textbf{23.51} & \textbf{0.859} & \textbf{0.106} & 120.4 & \textbf{0.586} \\
        Depth & 23.41 & 0.858 & \textbf{0.106} & \textbf{119.7} & 0.581 \\
        \bottomrule
    \end{tabular}%
    }
    \caption{\textbf{Ablation study on different visual action prompts.}
    Representations with more details (mesh, depth) perform marginally better. Skeleton is preferred for its acquisition efficiency.
    }
    \label{tab:ablation_visual_prompting}
\end{table}

\nbf{Model architecture}
We evaluate different modules' contributions by training models on the DROID~\cite{droid} dataset. As shown in~\cref{tab:ablation_architecture},~ControlNet~\cite{controlnet} plays a more critical role in both generation quality and dynamic accuracy. Injecting visual action prompts to the main DiT and utilize LoRA-based fine-tuning of the DiT backbone~\cite{lora,cogvideox} is also effective, which yields marginal gains.

\begin{table}[t]
    \centering
    \resizebox{0.96\columnwidth}{!}{%
    \begin{tabular}{lccccc}
        \toprule
        Model Variant & PSNR $\uparrow$ & SSIM $\uparrow$ & LPIPS $\downarrow$ & FVD $\downarrow$ & ST-IoU $\uparrow$ \\
        \midrule
        w/o ControlNet & 20.19 & 0.819 & 0.151 & 165.2 & 0.396 \\
        w/o Main Branch & 21.09 & 0.830 & 0.138 & 146.9 & 0.442 \\
        \textbf{Ours (full)} & \textbf{21.26} & \textbf{0.834} & \textbf{0.132} & \textbf{141.8} & \textbf{0.450} \\
        \bottomrule
    \end{tabular}%
    }
    \caption{\textbf{Ablation study on model architecture.}}
    \label{tab:ablation_architecture}
\end{table}

\section{Conclusion}
\label{sec:conclusion}

In this paper, we propose \shortName~as universal action representations for action-to-video generation which effectively represent complex high-DoF actions and retain the cross-domain dynamic transfer capability of video generation models at the same time. We design robust pipelines for building visual action prompts from heterogeneous data sources for training, and utilize lightweight fine-tuning to inject visual action prompts into a pretrained video generation model. Our method demonstrates improvements in both interaction fidelity and domain adaptability, with experimental results validating the model's effectiveness.

\nbf{Limitations and future works}
Our method still faces two main limitations. First, current visual action prompts represent actions in 2D, offering limited 3D cues. Integrating additional sparse 3D information could introduce better 3D awareness. Moreover, the base model is pre-trained on text-to-video tasks where motion is explicitly specified through texts, which hasn't been effectively utilized. Future works could adapt the attention between video-text tokens to video-action tokens, injecting action control more effectively.

\nbf{Acknowledgement}
This work was partially supported by the Major Program of Xiangjiang Laboratory (No. 24XJJCYJ01004), NSFC (NO. 62322207, NO. U24B20154), and Information Technology Center and State Key Lab of CAD\&CG, Zhejiang University.

{\small
\bibliographystyle{ieeenat_fullname}
\bibliography{sections/11_references}
}

\end{document}